\documentclass[runningheads]{llncs}
\usepackage{graphicx}

\usepackage{tikz}
\usepackage{comment}
\usepackage{amsmath,amssymb} 
\usepackage{color}
\usepackage{upgreek}
\usepackage{subcaption}
\usepackage{bbm}

\usepackage[accsupp]{axessibility}  

\begin{document}
\pagestyle{headings}
\mainmatter
\def\ECCVSubNumber{2856}  

\title{Robust Category-Level 6D Pose Estimation with Coarse-to-Fine Rendering of Neural Features} 


\titlerunning{Robust 6D Pose Estimation}
%
\author{Wufei Ma\inst{1} \and
Angtian Wang\inst{1} \and
Alan Yuille\inst{1} \and
Adam Kortylewski\inst{1,2,3}}
\authorrunning{W. Ma et al.}
%
\institute{Johns Hopkins University, Baltimore MD 21218, USA\\
\email{\{wma27,angtianwang,ayuille1,akortyl1\}@jhu.edu} \and
Max Planck Institute for Informatics, Saarbrücken, Germany\\
\and University of Freiburg, Germany}
\maketitle

\begin{abstract}
We consider the problem of category-level 6D pose estimation from a single RGB image. Our approach represents an object category as a cuboid mesh and learns a generative model of the neural feature activations at each mesh vertex to perform pose estimation through differentiable rendering. A common problem of rendering-based approaches is that they rely on bounding box proposals, which do not convey information about the 3D rotation of the object and are not reliable when objects are partially occluded. Instead, we introduce a coarse-to-fine optimization strategy that utilizes the rendering process to estimate a sparse set of 6D object proposals, which are subsequently refined with gradient-based optimization. The key to enabling the convergence of our approach is a neural feature representation that is trained to be scale- and rotation-invariant using contrastive learning. Our experiments demonstrate an enhanced category-level 6D pose estimation performance compared to prior work, particularly under strong partial occlusion.
\end{abstract}

\section{Introduction}
Estimating the 3D position and 3D orientation of objects is an important requirement for a comprehensive scene understanding in computer vision. Real-world applications, such as augmented reality (AR) or robotics, require vision systems to generalize in new environments that may contain previously unseen and partially occluded object instances.
However, most prior work on 6D pose estimation focused on the ``instance-level'' task, where exact CAD models of the object instances are available \cite{PoseCNN,PVNet,Li_2018_ECCV,PVN3D,RePOSE}. Moreover, the few prior methods on ``category-level'' 6D pose estimation often either rely on a ground truth depth map \cite{NOCS,Lin_2021_ICCV}, which are practically hard to obtain in many application areas, or rely on 2D bounding box proposals \cite{zhou2018starmap,wang2021nemo}, which are not reliable in challenging occlusion scenarios \cite{wang2020robust} (see also our experimental results).

Recent work introduced generative models of neural network features for image classification \cite{kortylewski2020compositional} and 3D pose estimation \cite{wang2021nemo}, which have the ability to learn category-level object models that are highly robust to partial occlusion. Intuitively, these models are composed of a convolutional neural network \cite{lecun1995convolutional} and a Bayesian generative model of the neural feature activations.
The invariance properties of the neural features enable these models to generalize despite variations in instance-specific details such as changes in the object shape and texture. 
Moreover, the generative model can be augmented with an outlier model \cite{huber2004robust} to avoid being distorted by local occlusion patterns.

\begin{figure}[t]
    \centering
    \includegraphics[width=0.95\textwidth]{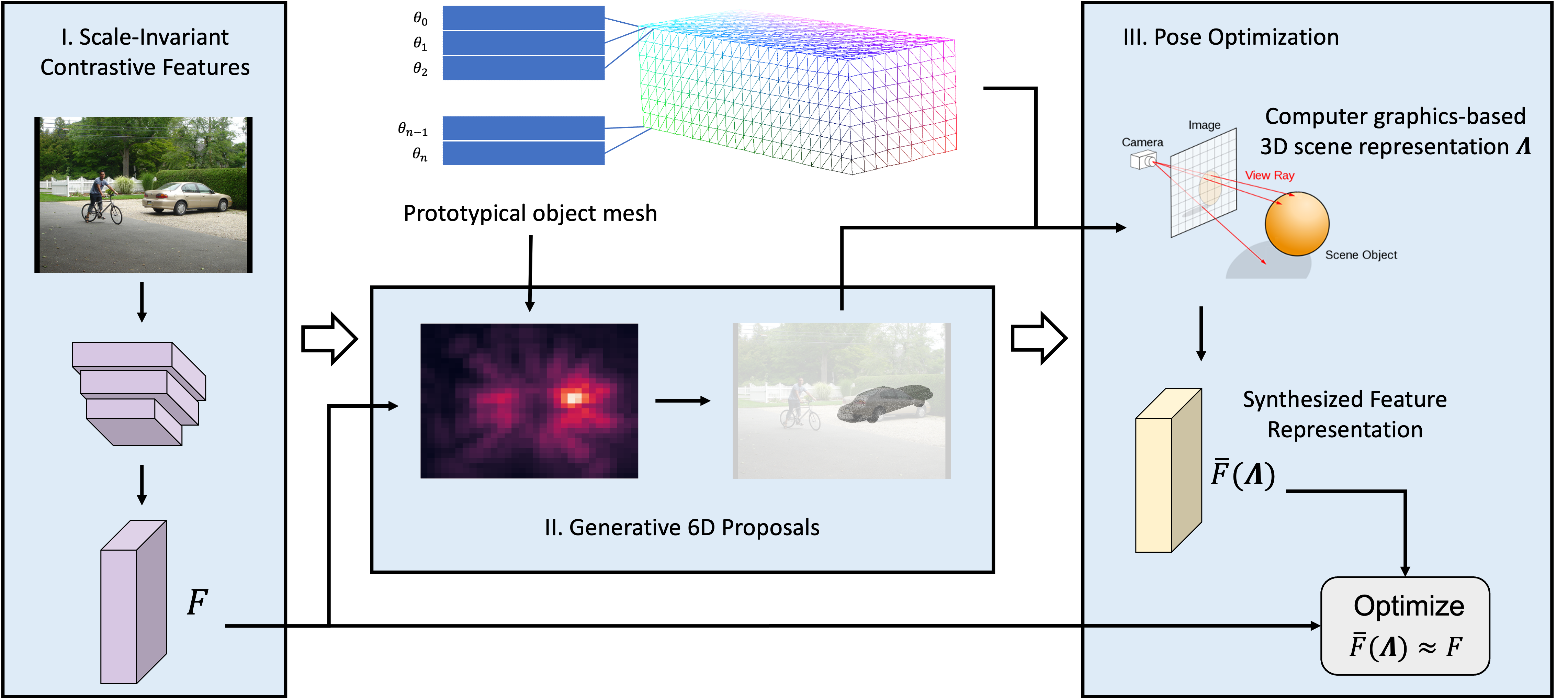}
    \caption{Overview of our coarse-to-fine 6D pose estimation. We propose to train a neural representation that is invariant to instance-specific details, 3D rotation, and changes in the object scale. With the help of the scale-invariant contrastive features, we can efficiently search generative 6D proposals in the coarse stage and then refine the initial 6D poses with pose optimization in the fine stage.}
    \label{fig:intro}
\end{figure}

In this work, we build on and significantly extend generative models of neural network features to perform category-level 6D pose estimation from a single RGB image.
In particular, we follow neural mesh models \cite{wang2021nemo} and represent an object category as a cuboid mesh and learn a generative model of the neural feature activations at each mesh vertex to perform pose estimation through a render-and-compare process. 
The core problem of such a rendering-based approach to pose estimation is to search efficiently through the combinatorially large space of the 6D latent parameters, because the iterative rendering process is rather costly compared to simple feed-forward regression approaches.
Related work addresses this problem by first estimating 2D object bounding boxes with a proposal network \cite{li2020rtm3d,NOCS,RePOSE}, but these are not reliable under partial occlusion and truncation \cite{wang2020robust}.
Instead, we address this problem by extending neural mesh models with scale-invariant features and a coarse-to-fine render-and-compare optimization strategy, which retains the robustness to partial occlusion (Figure \ref{fig:intro}).
In particular, we first use a coarse search strategy, in which we render the model in a set of pre-defined initial poses that are evenly distributed over the entire search space and select a sparse set of candidate initializations with low reconstruction loss.
This process generates 6D object proposals robustly under occlusion, as it relies on the generative object model. 
Subsequently, the 6D proposals are refined with gradient-based render-and-compare optimization. 

The key to making the coarse-to-fine optimization efficient and accurate (i.e., not missing out on small or partially occluded objects) is to learn a feature representation that induces large convergence basins in the optimization process.
To this end, we introduce a contrastive learning framework \cite{He_2020_CVPR,Wu_2018_CVPR,bai2020coke} to learn features that are invariant to instance-specific details (such as changes in the shape and texture), as well as to changes in the 3D pose and scale of the object. 

We evaluate our model on the PASCAL3D+ dataset \cite{xiang2014beyond} and the ObjectNet3D dataset \cite{xiang2016objectnet3d}, which are challenging real-world datasets of outdoor and indoor scenes, respectively. Our experiments demonstrate that our model outperforms strong object detection and pose estimation baseline models. Our model further demonstrates exceptional robustness to partial occlusion compared to all baseline methods on the Occluded PASCAL3D+ dataset \cite{wang2020robust}.

The main contributions of our work are:
\begin{enumerate}
\item We introduce a render-and-compare approach for category-level 6D pose estimation and adopt a coarse-to-fine pose estimation strategy that is accurate and highly robust to partial occlusion.
\item We use a contrastive learning framework to train a feature representation invariant to instance-specific details, 3D rotation, and changes in object scale.
\item The invariant features enable a coarse-to-fine render-and-compare optimization, which involves novel generative 6D object proposals and a subsequent gradient-based pose refinement. 
\item Our method outperforms previous methods on the PASCAL3D+, ObjectNet3D dataset, and we demonstrate the robustness of our model to partial occlusion on Occluded PASCAL3D+ dataset. We further show the efficacy of our proposed modules in the ablation study.
\end{enumerate}

\section{Related Work}

\textbf{Category-level 3D pose estimation.} Category-level 3D pose estimation assumes the bounding box of the object is given and predicts the 3D object pose. Previous methods can be categorized into two groups, keypoint-based methods and render-and-compare methods. Keypoint-based methods \cite{7989233,zhou2018starmap} first detect semantic keypoints and then predict 3D object pose by solving a Perspective-n-Point problem. Render-and-compare methods \cite{wang2021nemo} predict the 3D pose by fitting a rigid transformation of the mesh model that minimizes the reconstruction error between a predicted feature map and a rendered feature map. 3D pose estimation methods often exploit the inductive bias that the principal points of the objects are close to the image center and the objects have a similar scale.\\



\textbf{Category-level 6D pose estimation.} Category-level 6D pose estimation is a more challenging problem and involves object detection and pose estimation without knowing the accurate 3D model or the textures of the testing objects. Previous methods \cite{NOCS,10.1007/978-3-030-58574-7_9,Lin_2021_ICCV} often investigate this problem in the RGBD setting. Depth maps help these models to infer the 3D location of the objects and at the same time resolve the scale ambiguities \cite{10.1007/978-3-030-58574-7_9}. However, depth annotations are often hard to obtain, which limits the practicality of these methods. In this work, we investigate category-level 6D pose estimation from monocular RGB images and show that our method can robustly estimate 6D object poses under partial occlusion and truncation.

\textbf{Feature-level render-and-compare.} Render-and-compare methods minimize the reconstruction error between a predicted feature representation and a representation rendered from a 3D scene (e.g., a 3D mesh model $\mathcal{M}$ and the corresponding 3D pose $m$). Previous methods follow similar formulations but differ in the feature representation and the optimization algorithms. Wang et al. \cite{NOCS} proposed to hard-code the features as the normalized 3D coordinates and predict the object pose by solving a rigid transformation between the 3D model $\mathcal{M}$ and the predicted coordinate map with the Umeyama algorithm \cite{umeyama1991least}. NeMo \cite{wang2021nemo} learns contrastive features for the 3D model $\mathcal{M}$ and solves 3D object pose with the objects centered and rescaled. Iwase et al. \cite{RePOSE} found that features with only 3 channels are sufficient for instance-level 6D pose estimation and proposed to learn the features with a differentiable Levenberg-Marquardt (LM) optimization.

In a broader context, the feature-level render-and-compare process can be interpreted as an \textit{approximate} analysis-by-synthesis \cite{grenander1970unified,grenander1996elements} approach to computer vision. Analysis-by-synthesis has several advantages over purely discriminative methods as it enables 
efficient learning \cite{wang2021neural} and largely enhances robustness in out-of-distribution situations, particularly when objects are partially occluded in image classification
\cite{kortylewski2020compositionalijcv,kortylewski2020combining,Yuan_2021_CVPR,xiao2020tdmpnet}, object detection \cite{wang2020robust},
scene understanding \cite{abs_scene,abs_scene2}, face reconstruction \cite{egger2018occlusion} and human detection \cite{girshick2011object}, as well as when objects are viewed from unseen 3D poses \cite{wang2021nemo}. 
Our work extends the approximate analysis-by-synthesis approach to category-level 6D pose estimation.

\section{Method}
This section presents our main contributions. First, we review the render-and-compare approach for pose estimation in Section \ref{sec:prior-work}. Then we introduce the learning of scale-invariant contrastive features in Section \ref{sec:learn-features}. In Section \ref{sec:coarse-to-fine}, we introduce a coarse-to-fine optimization strategy that uses a generative model to generate 6D object proposals in the coarse stage and then refines the initial 6D poses with a render-and-compare pose optimization. 
We discuss a multi-object reasoning module in Section \ref{sec:multi-obj} that enables our model to accurately detect occluded and truncated objects and as well as complicated multi-object scenes.

\subsection{Notation}
We denote a feature representation of an input image $I$ as $\zeta(I) = F^l \in \mathbb{R}^{H\times W \times c}$. 
Where $l$ is the output of layer $l$ of a deep convolutional neural network $\zeta$, with $c$ being the number of channels in layer $l$. $f^l_i \in \mathbb{R}^c$ is a feature vector in $F^l$ at position $i$ on the 2D lattice $\mathcal{P}$ of the feature map.
In the remainder of this section, we omit the superscript $l$ for notational simplicity because this is fixed a-priori in our model.

\subsection{Prior Work: Render-And-Compare for Pose Estimation} \label{sec:prior-work}

Our work builds on and significantly extends neural mesh models (NMMs) \cite{RePOSE,wang2021nemo}, which are themselves 3D extensions of Compositional Generative Networks \cite{kortylewski2021compositional}. Neural mesh models define a probabilistic generative models $p(F \mid \mathfrak{N})$ of the real-valued feature activations $F$ using a 3D neural mesh representation $\mathfrak{N}$. The neural mesh $\mathfrak{N} = \{\mathcal{V}, \mathcal{E}, \mathcal{C}\}$ is represented by a set of vertices $\mathcal{V} = \{V_i \in \mathbb{R}^3\}_{i=1}^N$ and learnable features for each vertex $\mathcal{C} = \{C_i \in \mathbb{R}^c\}_{i=1}^N$, where $c$ is the number of channels in layer $l$. Given the object pose (or camera viewpoint) $m$, we can render the neural mesh model $\mathfrak{N}$ into feature maps using rasteriation, i.e., $\bar{F}(m) = \mathfrak{R}(\mathfrak{N}, m) \in \mathbb{R}^{H \times W \times D}$. The neural mesh model defines the likelihood of a target feature map $F \in \mathbb{R}^{H \times W \times D}$ as
\begin{align}
    p(F \mid \mathfrak{N}, m, \mathcal{B}) = \prod_{i \in \mathcal{FG}} p(f_i \mid \mathfrak{N}, m) \prod_{i' \in \mathcal{BG}} p(f_{i'} \mid \mathcal{B}) \label{eq:likelihood}
\end{align}
where the foreground $\mathcal{FG}$ is the set of all positions on the 2D lattice $\mathcal{P}$ of the feature map $F$ that are covered by the rendered neural mesh model and the background $\mathcal{BG}$ contains those pixels that are not covered by the mesh. The foreground likelihood is defined as a Gaussian distribution $p(f_i \mid \mathfrak{N}, m) = \mathcal{N}(f_i \mid C_r, \sigma_r^2I)$. The correspondence between the image feature $f_i$ and the vertex feature $C_r$ is determined through the rendering process. Background features are modeled using a simple background model that is defined by a Gaussian distribution $p(f_{i'} \mid \mathcal{B}) = \mathcal{N}(f_{i'} \mid b, \sigma^2 I)$ with $\mathcal{B} = \{b, \sigma\}$, which can be estimated with maximum likelihood from the background features.
The training of the generative model parameters $\{ \mathfrak{N}, B \}$ and the feature extractor is done by maximum likelihood estimation (MLE) from the training data. 
At test time, we can infer the object pose $m$ by minimizing the negative log-likelihood of the model w.r.t. the pose $m$ with gradient descent
\begin{align}
    \mathcal{L}_\text{NLL}(F, \mathfrak{N}, m, \mathcal{B}) = & - \ln p(F \mid \mathfrak{N}, m, \mathcal{B}) \nonumber \\
    = & - \sum_{i \in \mathcal{FG}} \left( \ln \left( \frac{1}{\sigma_r \sqrt{2\pi}} \right) - \frac{1}{2\sigma_r^2} \lVert f_i - C_r \rVert^2 \right) \nonumber \\
    & - \sum_{i' \in \mathcal{BG}} \left( \ln \left( \frac{1}{\sigma \sqrt{2 \pi}} \right) - \frac{1}{2\sigma^2} \lVert f_{i'} - b\rVert^2 \right) \label{eq:neg-log-likelihood}
\end{align}
Assuming unit variance \cite{wang2021nemo}, i.e., $\sigma_r = \sigma = 1$, the loss function reduce to the mean squared error (MSE) between vertex features and the target feature map
\begin{align}
    \mathcal{L}_\text{NLL}(F, \mathfrak{N}, m, \mathcal{B}) = \frac{1}{2} \sum_{i \in \mathcal{FG}} \lVert f_i - C_r \rVert^2 + \frac{1}{2} \sum_{i' \in \mathcal{BG}} \lVert f_i - b\rVert^2 + \textit{const.} \label{eq:nll}
\end{align}

Previous works adopted this general framework for category-level 3D pose estimation \cite{wang2021nemo,NOCS} and instance-level 6D pose estimation \cite{RePOSE}, thereby using different types of learnable features $\mathcal{C}$ and optimization algorithms.
In this work, we extend this framework to category-level 6D pose estimation from a single RGB image, which requires us to overcome additional challenges. 
Specifically, we need to address the challenge that the learnable feature representation $\mathcal{C}$ needs to account for the large variations in object scale, as well as the intra-category variation in terms of the object shape and texture properties.

\subsection{Learning Scale-Invariant Contrastive Features} \label{sec:learn-features}

\begin{figure}[t]
\centering
\begin{subfigure}[b]{0.22\textwidth}
\includegraphics[width=\textwidth]{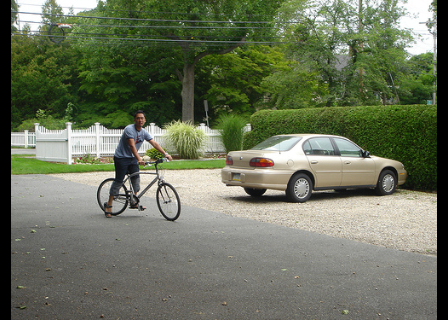}
\caption{}
\label{fig:converge-a}
\end{subfigure}
\hfill
\begin{subfigure}[b]{0.22\textwidth}
\includegraphics[width=\textwidth]{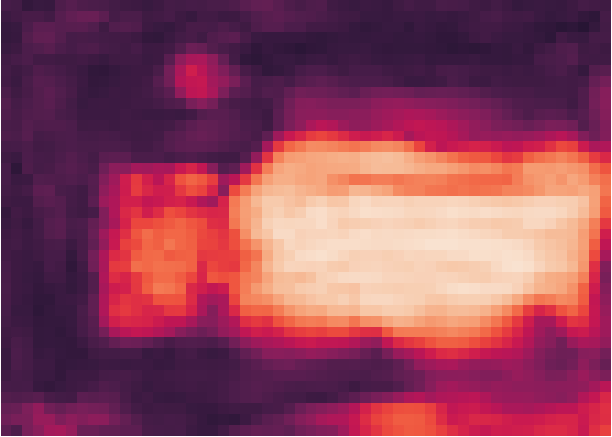}
\caption{}
\label{fig:converge-b}
\end{subfigure}
\hfill
\begin{subfigure}[b]{0.22\textwidth}
\includegraphics[width=\textwidth]{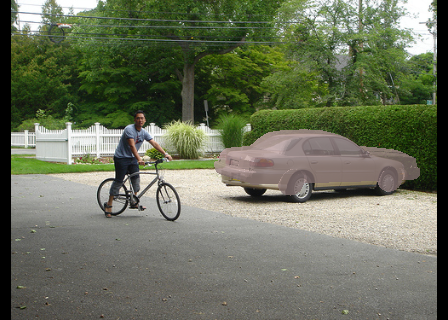}
\caption{}
\label{fig:converge-c}
\end{subfigure}
\begin{subfigure}[b]{0.31\textwidth}
\includegraphics[width=\textwidth]{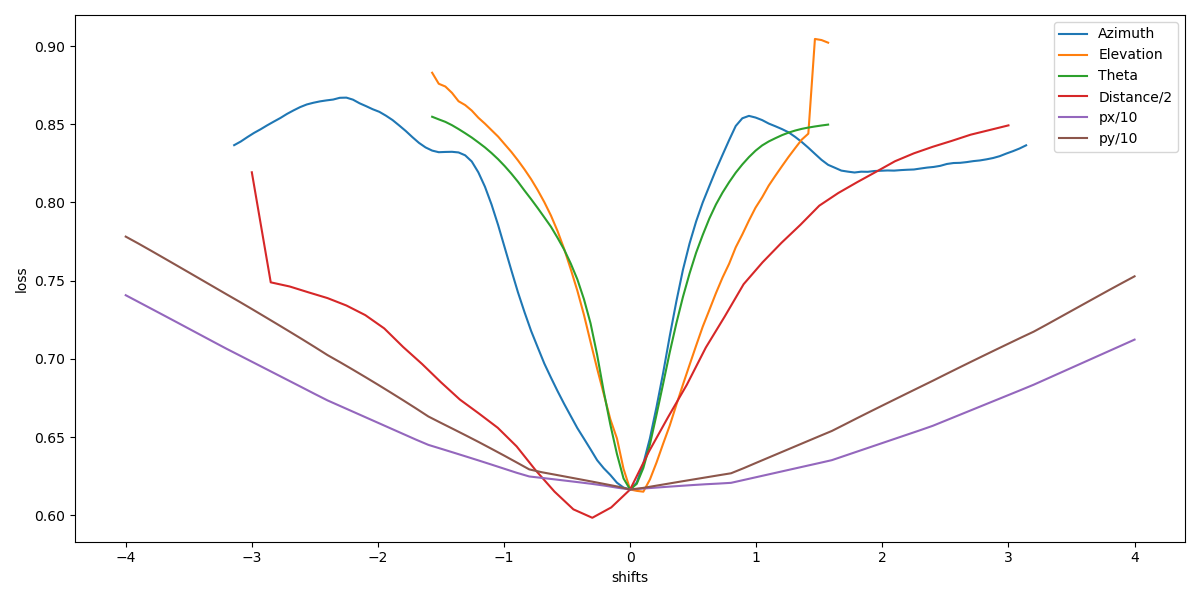}
\caption{}
\label{fig:converge-d}
\end{subfigure}
\caption{We visualize the loss landscapes of the pose optimization with scale-invariant contrastive features. (a) shows the input image. (b) shows the reconstruction loss $\max_r \lVert C_r - f_i \rVert^2$ for each pixel. (c) visualizes the predicted 6D pose. Each curve in (d) corresponds to one of the six dimensions of the 6D pose and is centered at the ground truth pose. We can see with the help the scale-invariant contrastive features, the pose optimization has a clear global minimum near the ground truth pose and is easy to optimize. This further allows us to search for generative 6D proposals, as described in Section \ref{sec:coarse-to-fine}.}
\label{fig:converge}
\end{figure}

In this work, we propose to account for the variations in the object scale, shape, and appearance by learning contrastive features that are invariant to these variations. This will enable us to estimate the 6D object pose by optimizing the maximum likelihood formulation in Equation \ref{eq:nll} directly with gradient-based optimization. 
We demonstrate the efficacy of our scale-invariant contrastive features in Figure \ref{fig:converge} and quantitatively in Section \ref{sec:quant}.

\textbf{Contrastive Learning of Scale-Invariant Features.} One of the major challenges in 6D pose estimation is the variation in object scales. Due to the nature of convolution layers in the feature extractor $\zeta$, nearby and distant objects could yield very different feature activations in $F$. Unfortunately, annotations of 6D poses for small objects are limited. Therefore, we use data augmentation to learn scale-invariant features from object-centric samples.\\

\begin{figure}[t]
\centering
\includegraphics[width=0.9\textwidth]{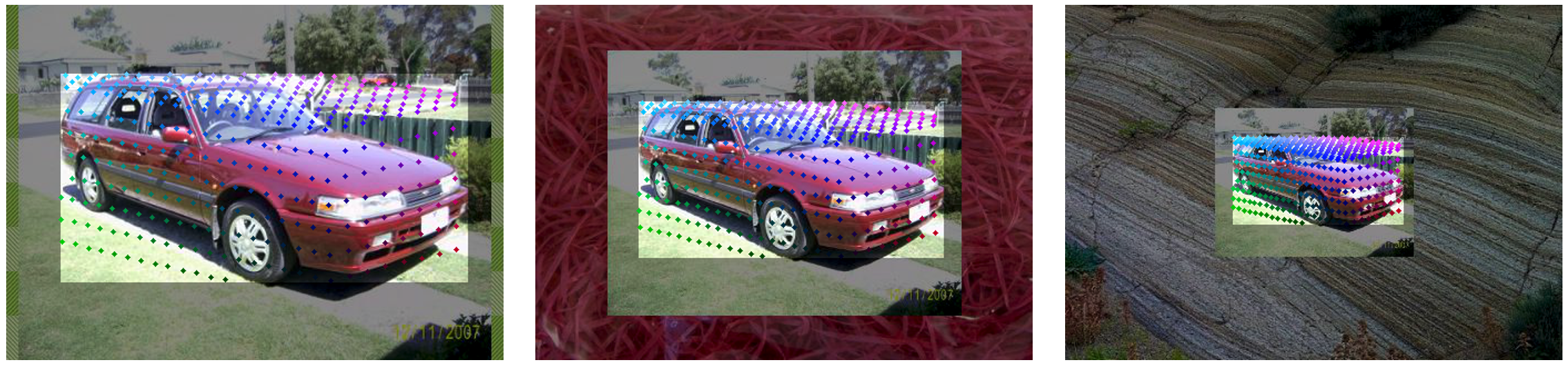}
\caption{Illustration of our object-centric data augmentation strategy, to generate feature activations across several scales, which are essential for our contrastive learning framework.}
\label{fig:object-centric}
\end{figure}

Specifically, given an image $\mathbf{I} \in \mathbb{R} ^ {H \times W \times 3}$, we prepare the training sample as follows. First, we resize the image with scale $s$ and obtain a new image with size $\frac{H}{s} \times \frac{W}{s}$. Then texture images from the Describable Textures Dataset (DTD) \cite{cimpoi14describing} are used to pad the image back to $H \times W$. We update the distance annotation $d$ of the object assuming a pinhole camera model, such that the distance annotation can be computed as $d' = d \cdot s$. The augmented data is depicted in Figure \ref{fig:object-centric}.

In order for the CNN backbone $\zeta$ to extract feature invariant to instance-specific details and to avoid local optima in the loss landscapes of the reconstruction loss, we train the feature extractor $\zeta$ using contrastive learning to learn features that are distributed akin to the probabilistic generative model as defined in Equations \ref{eq:likelihood}-\ref{eq:nll}. We achieve this by adopting a contrastive loss:
\begin{align}
    \mathcal{L}_\text{contrastive} = - \sum_{i \in \mathcal{FG}} \sum_{j \in \mathcal{FG} \setminus \{i\}} \lVert f_i - f_j \rVert^2 - \sum_{i \in \mathcal{FG}} \sum_{j \in \mathcal{BG}} \lVert f_i - f_j \rVert^2
\end{align}
which encourages the features of different vertices to be discriminative from each other and features of the object vertices distinct from the features in the background. Our full model is trained by optimizing $\mathcal{L}_\text{contrastive}$ in a contrastive learning framework, where we update the parameters of the feature extractor $\zeta$ and the vertices features $\mathcal{C}$ in the neural mesh model jointly.


\textbf{MLE Learning of the Neural Mesh Model (NMM).} We train the parameters $\mathcal{C}$ of the probabilistic generative model through maximum likelihood estimation (MLE) by minimizing the negative log-likelihood of the feature representations over the whole training set (Equation \ref{eq:nll}). The correspondence between the feature vectors $f_i$ and vertices $r$ is computed using the annotated 6D pose. To reduce the computational cost of optimizing Equation \ref{eq:nll}, we follow \cite{bai2020coke} and update $\mathcal{C}$ in a moving average manner.\\

\textbf{Convergence properties.} The benefits of the scale-invariant contrastive features are two-fold. 
First, the ground truth 6D pose is very close to the global minimum of the reconstruction loss in all six dimensions.
We illustrate the 6D loss landscapes in Figure \ref{fig:converge}. 
Each curve corresponds to one of the six dimensions of the 6D pose and is centered at the ground truth pose.
The large convergence basins that can be observed allow us to search for object proposals from simply sparse sampling and to evaluate a pre-defined set of 6D poses, without the need of a first-stage model widely used by related works \cite{li2020rtm3d,NOCS,RePOSE}. 
Second, the loss landscapes are generally smooth around the global minimum. This contrasts with the keypoint-based methods that fit a rigid transformation between two groups of keypoints \cite{li2020rtm3d,NOCS} and the render-and-compare methods over RGB space \cite{10.1145/311535.311556,schonborn2017markov} with many local minima on the optimization surface. \\

\subsection{Coarse-to-Fine 6D Pose Estimation} \label{sec:coarse-to-fine}


Previous methods for 3D object detection or 6D pose optimization are built on top of a 2D region proposal network or refine predictions from a separate pose estimation network. Although this approach was empirically effective, the performance of the hybrid model is largely limited by the 2D region proposal network or the initial pose estimation network. The first-stage networks are unreliable for objects with out-of-distribution textures or shapes, or even miss the object if the object is partially occluded or truncated.

\begin{figure}[t]
\centering
\includegraphics[width=0.85\textwidth]{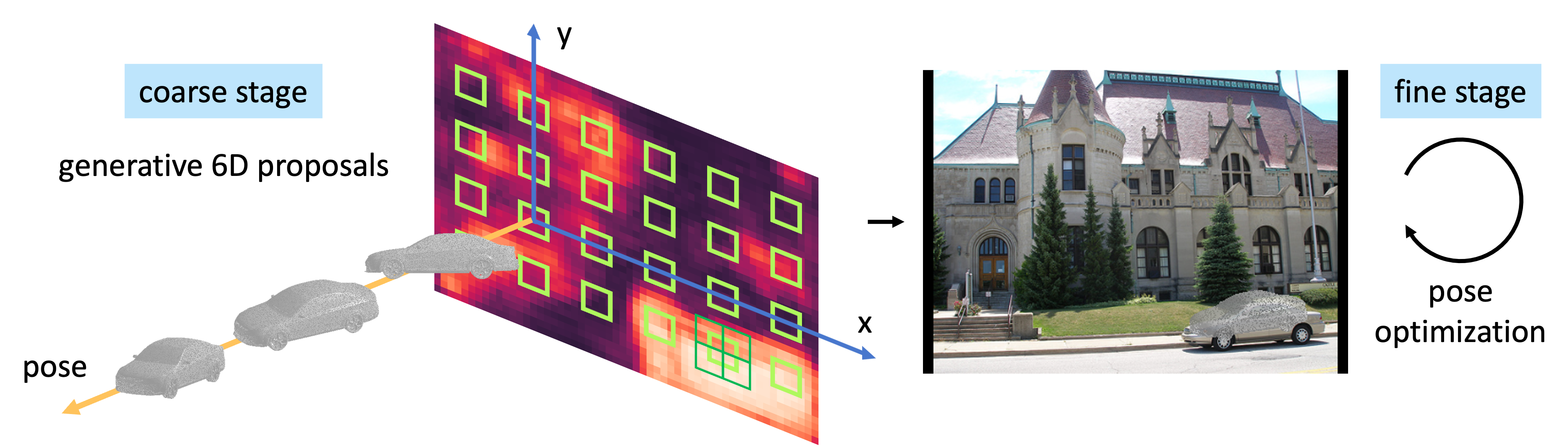}
\caption{Overview of our proposed coarse-to-fine 6D pose estimation. We search for generative 6D proposals in the coarse stage and then refine the initial 6D poses with pose optimization in the fine stage. Since the generative 6D proposals are built on the generative neural mesh models and the scale-invariant contrastive features, they are robust to partial occlusion and truncation and are easy to optimize. Note that the CAD models are for visualization only and are not used in any part of our model.}
\label{fig:coarse-to-fine}
\end{figure}

Therefore, we propose a coarse-to-fine 6D pose estimation strategy that searches generative 6D proposals in the coarse stage and then refines the initial 6D poses with pose optimization in the fine stage. The overview of our coarse-to-fine strategy is depicted in Figure \ref{fig:coarse-to-fine}. Since the generative 6D proposals are built on the generative neural mesh models and scale-invariant contrastive features, they are robust to partial occlusion and truncation. Moreover, this coarse-to-fine strategy can largely benefit subsequent pose optimization. The generative 6D proposals are often located at regions near global optimum that makes effective gradients toward the ground truth pose. We compare Faster R-CNN 6D proposals and our generative 6D proposals quantitatively in Section \ref{sec:ablation}. We further visualize loss landscapes of different 6D object proposals in the supplementary material.

\textbf{Coarse stage: generative 6D proposals.} With the scale-invariant contrastive features, the pose optimization has a clear global minimum near the ground truth location, and the loss landscapes are smooth with decent gradients around the global minimum (see Figure \ref{fig:converge}). This nice property allows us to search for generative 6D proposals from a sparse sampling over six dimensions. Given a 6D pose sample, we estimate the reconstruction loss in Equation \ref{eq:nll} and predict generative 6D proposals with non-maximum suppression. Since the structure of the 3D model $\mathcal{M}$ and the sampled 6D poses are consistent across all testing samples, the 2D coordinates and visibility of the vertices can be pre-computed and cached. We further adopt a strategy to speed up by searching over the 2D locations first and then the other four dimensions, which is detailed in the supplementary materials. This allows us to predict generative 6D proposals that are robust to partial occlusion and truncation and are easy to optimize with a negligible complexity overhead.\\


\textbf{Fine stage: pose optimization.} The goal of pose optimization is to refine the initial 6D pose of an object, which can be either predicted by a stand-alone pose estimation network or from our generative 6D proposals. We maximize the feature correlation between the predicted features $F$ from the feature extract $\zeta$ and the rendered features $\bar{F}$ with respect to the object pose. Since the ground truth $\mathcal{FG}$ and $\mathcal{BG}$ is unknown, we approximate the maximum likelihood problem in Equation \ref{eq:nll} with a one-hot map $\mathbf{Z}$ to denote the foreground and background regions in the feature map:
\begin{align}
    \mathbf{Z}_i = \begin{cases}
    1 & \text{if $\lVert f_i - C_r \rVert^2 \geq \lVert f_i - b \rVert^2$} \\ 0 & \text{otherwise}
    \end{cases}
\end{align}
Finally, we minimize $\mathcal{L}_\text{NLL}$ with respect to object pose $m$ with gradient descent
\begin{align}
    \mathcal{L}_\text{NLL}(F, \mathfrak{N}, m, \mathcal{B}) = \frac{1}{2} \sum_i \left( \mathbbm{1}_{Z_i = 1} \lVert f_i - C_r \rVert^2 + \mathbbm{1}_{Z_i = 0} \lVert f_i - b\rVert^2 \right) + \textit{const.}
\end{align}

\subsection{Multi-Object Reasoning} \label{sec:multi-obj}

One challenge when extending 3D pose estimation to 6D pose estimation is the existence of multiple objects in the image. Therefore, we propose a multi-object reasoning module that can resolve mutual occlusion and can be applied on top of any render-and-compare methods. The motivation is that we need to assign the pixels in the feature maps to different instances, and our multi-object reasoning module resembles related methods in instance segmentation \cite{Yuan_2021_CVPR}.

Given multiple generative 6D proposals, we run the pose optimization gradient descent for a small number of epochs. If the rendered feature maps of two objects overlap, we recover the occlusion order by running pixel-level competition, and for each overlapping region, only one object is considered as the foreground object and the other objects are considered as background. We use a one-hot map to record the multi-object reasoning results, where
\begin{align}
    \mathbf{Z} \in \mathbb{Z}^{H \times W \times k}, \;\;\; Z_{i, j, k} = \begin{cases}
    1 & \text{if the $k$-th object is the foreground object} \\ 0 & \text{otherwise}
    \end{cases}
\end{align}
Then we run the pose optimization again given the occlusion ordering $\mathbf{Z}$. We visualize the results of the multi-object occlusion reasoning in Figure \ref{fig:qualitative}.


\section{Experiments}
In this section, we investigate the performance of our approach in challenging 6D pose estimation datasets and compare its performance to related methods. We first describe the experimental setup in Section \ref{sec:exp-setup}. Then we study the performance of our model in Section \ref{sec:quant}. We visualize some qualitative examples in Section \ref{sec:qualitative}. Finally, we run ablation study experiments on the generative 6D proposals and the multi-object reasoning module in Section \ref{sec:ablation}.

\subsection{Experimental Setup} \label{sec:exp-setup}

\textbf{Datasets.} We evaluate our model on PASCAL3D+ dataset \cite{xiang2014beyond}, Occluded PASCAL3D+ dataset \cite{wang2020robust}, and ObjectNet3D dataset \cite{xiang2016objectnet3d}. PASCAL3D+ dataset contains objects from 12 man-made categories, and each object is annotated with 3D pose, 2D centroid, and object distance. The ImageNet subset of the PASCAL3D+ dataset contains 11045 images for training and 10812 images for evaluation, and the PASCAL VOC subset contains 4293 images for training and 4212 images for validation. Occluded PASCAL3D+ is based on the ImageNet subset of the PASCAL3D+ dataset, and partial occlusion is simulated by superimposing occluders on top of the objects and the background. We also experimented on ObjectNet3D dataset, which consists of 100 categories with 17101 training images and 19604 testing images. Following \cite{zhou2018starmap,wang2021nemo}, we compare the 6D pose estimation performance on 18 categories.\\

\textbf{Evaluation metrics.} Category-level 6D pose estimation estimates both the 3D pose (azimuth, elevation, and in-plane rotation) and the 3D location of the visible objects. In our experiments, we adopt the pose estimation error and the average distance metric (ADD) for evaluation. Following \cite{zhou2018starmap}, the pose estimation error measures the angle between the predicted rotation matrix and the ground truth rotation matrix $\Delta(R_\text{pred}, R_\text{gt}) = \frac{\lVert \operatorname{logm} (R_\text{pred}^\top R_\text{gt}) \rVert_\mathcal{F}}{\sqrt{2}}$. Average distance (ADD) is a widely used metric to measure the translation of the keypoints between the ground truth pose and the predicted pose. For the PASCAL VOC images, we also evaluate the mean average precision (mAP) at $(\pi/3, 5.0)$.\\


\textbf{Implementation details.} Our model includes a contrastive feature backbone and a corresponding neural mesh model. The feature extractor is a ResNet50 model with two upsampling layers, so the output feature map is $\frac{1}{8}$ of the input resolution. The neural mesh model is a category-wise cuboid model with around 1100 vertices. The scale of the cuboid mesh model is the average of the scales of the sub-category mesh models, and the vertices are sampled uniformly across six faces. Our model is trained for 1200 epochs with random horizontal flip and 2D translation and takes around 20 hours on one NVIDIA RTX Titan GPU. During inference, the pose optimization with multi-object reasoning takes 4.1 seconds on average per object.\\

\textbf{Baseline models.} Since we know of no other 6D pose estimation methods for category-level 6D pose estimation from a single RGB image, we compare our model with related works in 3D object detection and 3D pose estimation and extend them to the 6D pose estimation setting.

RTM3D is one of the state-of-the-art models for monocular 3D object detection. It predicts a 3D bounding box (i.e., location, rotation, and scale) by minimizing the reprojection error between the regressed 2D keypoints and the corners of the 3D cuboid. To extend RTM3D to 6D pose estimation, we fix the cuboid dimensions and fit a rigid 6D transformation.\\

We further compare our approach with two-stage models that predict object proposals in the first stage and then estimate object poses from the proposed RoIs. We adopt Faster R-CNN \cite{ren2016faster} for object detection. Two methods are considered for pose estimation. Following previous works \cite{zhou2018starmap,wang2021nemo}, we formulate the pose estimation as a classification problem and predict the object pose from the RoI features from the Faster R-CNN backbone. Based on the reported results \cite{zhou2018starmap,wang2021nemo}, we also consider the state-of-the-art 3D pose estimation model, NeMo \cite{wang2021nemo}, where we optimize the 3D object pose from the predicted 2D bounding box. The two models are denoted as ``FRCNN+Cls'' and ``FRCNN+NeMo'' respectively.

\setlength{\tabcolsep}{4pt}
\begin{table}[t]\begin{center}
\caption{Quantitative results of 6D pose estimation on PASCAL3D+ dataset.}
\label{table:quant-pascal3dp}
\resizebox{\textwidth}{!}{%
\begin{tabular}{lcccccr}
\hline\noalign{\smallskip}
Subset & Model & Pose Acc ($\frac{\pi}{6}$) $\uparrow$ & Pose Acc ($\frac{\pi}{18}$) $\uparrow$ & Median Pose Error $\downarrow$ & Median ADD $\downarrow$ & mAP $\uparrow$ \\
\noalign{\smallskip}
\hline
\noalign{\smallskip}
ImageNet & FRCNN+Cls & 78.90 & 37.35 & 0.22 & 0.74 & - \\
ImageNet & FRCNN+NeMo & 66.06 & 28.44 & 0.33 & 1.84 & - \\
ImageNet & RTM3DExt & 74.94 & 39.56 & 0.23 & 0.92 & - \\
ImageNet & Ours & \textbf{81.45} & \textbf{47.68} & \textbf{0.19} & \textbf{0.53} & - \\
\noalign{\smallskip}
\hline
\noalign{\smallskip}
PASCAL VOC & FRCNN+Cls & 38.98 & 15.05 & 1.38 & 2.04 & 0.11 \\
PASCAL VOC & FRCNN+NeMo & 40.13 & \textbf{19.17} & 1.40 & 2.14 & 0.32 \\
PASCAL VOC & RTM3DExt & 18.04 & 8.12 & 6.28 & 20.0 & 0.11 \\
PASCAL VOC & Ours & \textbf{45.32} & 18.09 & \textbf{0.65} & \textbf{1.87} & \textbf{0.43} \\
\noalign{\smallskip}
\hline
\end{tabular}}
\end{center}
\end{table}
\setlength{\tabcolsep}{1.4pt}

\setlength{\tabcolsep}{4pt}
\begin{table}[t]\begin{center}
\caption{Quantitative results of 6D pose estimation on ObjectNet3D dataset.}
\label{table:quant-objectnet3d}
\resizebox{\textwidth}{!}{%
\begin{tabular}{lcccccccr}
\hline\noalign{\smallskip}
Model & Pose Acc ($\frac{\pi}{6}$) $\uparrow$ & Pose Acc ($\frac{\pi}{18}$) $\uparrow$ & Median Pose Error $\downarrow$ & Median ADD $\downarrow$ \\
\noalign{\smallskip}
\hline
\noalign{\smallskip}
RTM3DExt & 38.44 & 16.61 & 2.50 & 4.87 \\
Ours & \textbf{52.47} & \textbf{16.65} & \textbf{0.49} & \textbf{1.95} \\
\noalign{\smallskip}
\hline
\end{tabular}}
\end{center}\end{table}
\setlength{\tabcolsep}{1.4pt}

\subsection{Quantitative Results} \label{sec:quant}


\textbf{6D pose estimation on PASCAL3D+ and ObjectNet3D dataset.} Table \ref{table:quant-pascal3dp} shows the 6D pose estimation results on the ImageNet and PASCAL VOC subsets of the PASCAL3D+ dataset. Compared to the ImageNet images, the PASCAL VOC subset is more challenging as there are multiple objects with occlusion, truncation, as well as a larger variance in the object scale and location. Our model outperforms all baseline models in both the pose error and the average distance metric. To show our model can be applied to different man-made in-door and out-door categories, we also experiment on the ObjectNet3D dataset, and the results are shown in Table \ref{table:quant-objectnet3d}. Despite the considerable number of occluded and truncated images in ObjectNet3D dataset, our model achieves reasonable accuracy and outperforms the competitive baseline by a wide margin.

\textbf{Robust 6D pose estimation on the Occluded PASCAL3D+ dataset.}

In order to investigate the robustness under occlusion, we further evaluate each model on Occluded PASCAL3D+ dataset under different occlusion levels. The quantitative results are reported in Table \ref{table:quant-occ-pascal3dp}. As we can see, our model achieves superior performance across all occlusion levels and shows a wider performance gap compared to the performance on the un-occluded images.

\setlength{\tabcolsep}{4pt}
\begin{table}\begin{center}
\caption{Quantitative results of 6D pose estimation on the Occluded PASCAL3D+ dataset.}
\label{table:quant-occ-pascal3dp}
\resizebox{\textwidth}{!}{%
\begin{tabular}{lcccccr}
\hline\noalign{\smallskip}
Subset & Level & Method & Pose Acc ($\frac{\pi}{6}$) $\uparrow$ & Pose Acc ($\frac{\pi}{18}$) $\uparrow$ & Median Pose Error $\downarrow$ & Median ADD $\downarrow$ \\
\noalign{\smallskip}
\hline
\noalign{\smallskip}
ImageNet & 1 & FRCNN+Cls & 61.48 & 26.11 & 0.33 & 1.07 \\
ImageNet & 1 & FRCNN+NeMo & 48.34 & 17.46 & 0.55 & 1.90 \\
ImageNet & 1 & RTM3DExt & 43.55 & 17.68 & 0.82 & 3.29 \\
ImageNet & 1 & Ours & \textbf{66.63} & \textbf{30.84} & \textbf{0.31} & \textbf{0.77} \\
\noalign{\smallskip}
\hline
\noalign{\smallskip}
ImageNet & 2 & FRCNN+Cls & 41.95 & 14.75 & 0.75 & 1.47 \\
ImageNet & 2 & FRCNN+NeMo & 34.33 & 9.65 & 1.05 & 2.03 \\
ImageNet & 2 & RTM3DExt & 21.27 & 7.24 & 3.14 & 5.00 \\
ImageNet & 2 & Ours & \textbf{47.95} & \textbf{16.25} & \textbf{0.56} & \textbf{1.22} \\
\noalign{\smallskip}
\hline
\noalign{\smallskip}
ImageNet & 3 & FRCNN+Cls & 22.42 & \textbf{5.58} & 2.01 & 1.95 \\
ImageNet & 3 & FRCNN+NeMo & 18.19 & 3.32 & 2.40 & 2.35 \\
ImageNet & 3 & RTM3DExt & 10.17 & 3.11 & 3.14 & 19.92 \\
ImageNet & 3 & Ours & \textbf{27.43} & 5.30 & \textbf{1.07} & \textbf{1.94} \\
\noalign{\smallskip}
\hline
\end{tabular}}
\end{center}\end{table}
\setlength{\tabcolsep}{1.4pt}


\subsection{Qualitative Examples} \label{sec:qualitative}

\begin{figure}
\centering
\includegraphics[width=0.9\textwidth]{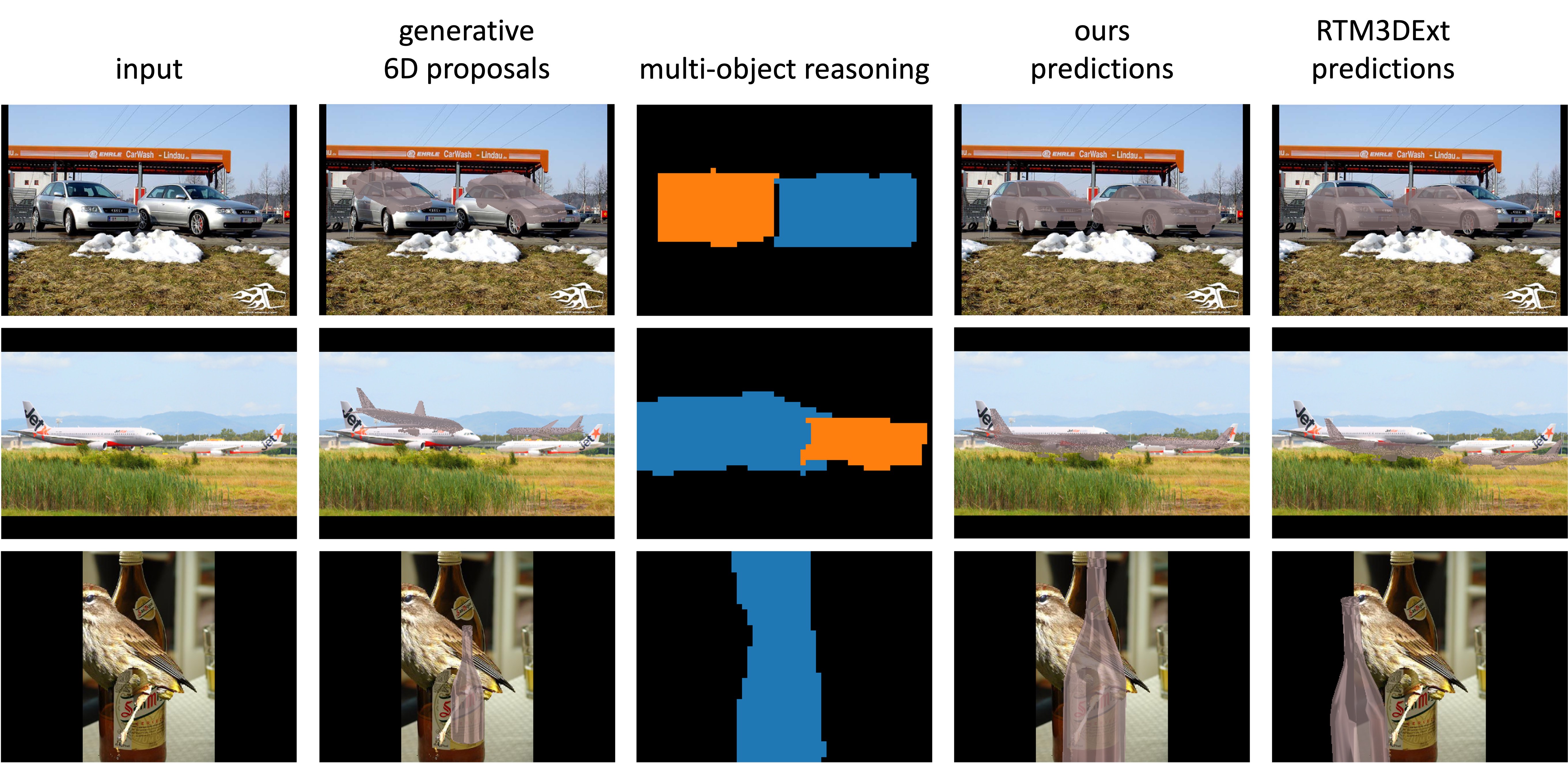}
\caption{Qualitative examples of our proposed model.}
\label{fig:qualitative}
\end{figure}

Figure \ref{fig:qualitative} shows some qualitative examples of our proposed model on PASCAL3D+ dataset. As we can see, our method can robustly estimate 6D poses for objects varying in scales and textures and is robust to partial occlusion.

\subsection{Ablation Study} \label{sec:ablation}

\textbf{Generative 6D proposals.} Unlike previous works that are based on 2D region proposal networks, we introduce generative 6D proposals that are robust to partial occlusion and truncation and are easy to optimize. We run ablation study experiments on the object proposal methods and compare the performance of our model using (1) generative 6D proposals (``Ours w/ GP''), or (ii) Faster R-CNN object proposals (``Ours w/ FRCNN''). The quantitative results on the Occluded PASCAL3D+ dataset are reported in Table \ref{table:ablation-6d-proposal}. As we can see, we can significantly improve the performance of our model in almost all tests with the generative 6D proposals. Note that ``Ours w/ FRCNN'' naturally benefits from model ensembling, and ``Ours w/ FRCNN'' can achieve a better pose estimation only in occlusion level 1 when the Faster R-CNN model can predict highly accurate poses without any refinement. In occlusion levels 2 and 3, ``Ours w/ GP'' significantly outperforms ``Ours w/ FRCNN'' in terms of both object location and pose estimation.

\setlength{\tabcolsep}{4pt}
\begin{table}\begin{center}
\caption{Ablation study on the 6D proposal method. We compare the performance of our proposed model using Faster R-CNN 6D proposals and generative 6D proposals (GP) on the PASCAL3D+ dataset.}
\label{table:ablation-6d-proposal}
\resizebox{\textwidth}{!}{%
\begin{tabular}{lccccr}
\hline\noalign{\smallskip}
Level & Method & Pose Acc ($\frac{\pi}{6}$) $\uparrow$ & Pose Acc ($\frac{\pi}{18}$) $\uparrow$ & Median Pose Error $\downarrow$ & Median ADD $\downarrow$ \\
\noalign{\smallskip}
\hline
\noalign{\smallskip}
1 & Ours w/ FRCNN & 65.79 & \textbf{34.56} & \textbf{0.28} & 0.95 \\
1 & Ours w/ GP & \textbf{66.63} & 30.84 & 0.31 & \textbf{0.77} \\
\noalign{\smallskip}
\hline
\noalign{\smallskip}
2 & Ours w/ FRCNN & 45.43 & \textbf{17.46} & 0.64 & 1.30 \\
2 & Ours w/ GP & \textbf{47.95} & 16.25 & \textbf{0.56} & \textbf{1.22} \\
\noalign{\smallskip}
\hline
\noalign{\smallskip}
3 & Ours w/ FRCNN & 23.53 & 5.27 & 1.67 & 2.05 \\
3 & Ours w/ GP & \textbf{27.43} & \textbf{5.30} & \textbf{1.07} & \textbf{1.94} \\
\noalign{\smallskip}
\hline
\end{tabular}}
\end{center}
\end{table}
\setlength{\tabcolsep}{1.4pt}

\textbf{Number of pre-defined initial poses.} Far all categories, we uniformly sample a sparse set of initial poses over the space of 6D poses. In Table \ref{table:ablation-num-initial-poses}, we ablate on the number of initial poses used to search the generative 6D proposals.

\setlength{\tabcolsep}{4pt}
\begin{table}\begin{center}
\caption{Ablation on the number of initial poses used to search the generative 6D proposals. The default setting in the paper is bold.}
\resizebox{\columnwidth}{!}{%
\begin{tabular}{lccccr}
\hline\noalign{\smallskip}
Init. 3D Pose & Init. 3D Loc & Acc ($\frac{\pi}{6}$) & Acc ($\frac{\pi}{18}$) & Median Err & Median ADD \\
\noalign{\smallskip}
\hline
\noalign{\smallskip}
12x3x3 & 3x3x3 & 69.77 & 35.82 & 0.28 & 0.65 \\
6x2x2 & 9x9x9 & 71.64 & 38.75 & 0.26 & 0.63 \\
\noalign{\smallskip}
\hline
\noalign{\smallskip}
\textbf{12x3x3} & \textbf{9x9x9} & 81.45 & 47.68 & 0.19 & 0.53 \\
\noalign{\smallskip}
\hline
\noalign{\smallskip}
18x6x6 & 9x9x9 & 83.46 & 49.28 & 0.18 & 0.51 \\
12x3x3 & 12x12x12 & 84.54 & 51.39 & 0.17 & 0.49 \\
\noalign{\smallskip}
\hline
\end{tabular}}
\label{table:ablation-num-initial-poses}
\end{center}
\end{table}

\setlength{\tabcolsep}{4pt}
\begin{table}
\begin{center}
\caption{Ablation study on the multi-object reasoning. We compare the performance of our proposed model with and without the multi-object reasoning on the PASCAL VOC subset of PASCAL3D+ dataset.}
\label{table:ablation-multi-object}
\resizebox{\textwidth}{!}{%
\begin{tabular}{lccccr}
\hline\noalign{\smallskip}
Method & Pose Acc ($\frac{\pi}{6}$) $\uparrow$ & Pose Acc ($\frac{\pi}{18}$) $\uparrow$ & Median Pose Error $\downarrow$ & Median ADD $\downarrow$ & mAP $\uparrow$ \\
\noalign{\smallskip}
\hline
\noalign{\smallskip}
Ours w/o reasoning & 44.89 & 17.63 & 0.66 & \textbf{1.87} & 0.41 \\
Ours & \textbf{45.32} & \textbf{18.09} & \textbf{0.65} & \textbf{1.87} & \textbf{0.43} \\
\noalign{\smallskip}
\hline
\end{tabular}}
\end{center}\end{table}
\setlength{\tabcolsep}{1.4pt}

\textbf{Multi-object reasoning.} In order to estimate 6D poses of multiple objects with render-and-compare, we propose a multi-object reasoning module to correctly assign pixels in the feature map to each object proposal. We quantitatively compare the performance of our model with and without an object reasoning module on the PASCAL subset of the PASCAL3D+ dataset. As shown in Table \ref{table:ablation-multi-object}, we can effectively improve the performance of our model with the multi-object reasoning module.

\section{Conclusions}
In this work, we consider the problem of category-level 6D pose estimation from a single RGB image. We find that previous methods built on 2D region proposal networks are less robust to partial occlusion and truncation, and the predicted initial poses are harder to optimize. Therefore, we propose a coarse-to-fine 6D pose optimization strategy where we search generative 6D proposals in the coarse stage and then refine them with pose optimization in the second stage. Both stages of our coarse-to-fine 6D pose estimation are built on our scale-invariant contrastive features and are hence robust to partial occlusion and truncation. We demonstrate the superiority of our approach compared to related works on several challenging datasets.

\textbf{Acknowledgements.} AK acknowledges support via his Emmy Noether Research Group funded by the German Science Foundation (DFG) under Grant No. 468670075. AY acknowledges the Institute for Assured Autonomy at JHU with Grant IAA 80052272, ONR N00014-21-1-2812, NSF grant BCS-1827427.

\clearpage
%
%
\bibliographystyle{splncs04}

\end{document}